\pdfoutput=1
\documentclass[10pt,twocolumn,letterpaper]{article}
\usepackage[pagenumbers]{cvpr}
\usepackage{microtype}
\usepackage{multirow}
\usepackage{pifont}
\graphicspath{{figures/}}
\newcommand{\method}{mmTherm}

\newcommand{\figref}[1]{Fig.~\ref{#1}}
\newcommand{\secref}[1]{Sec.~\ref{#1}}
\definecolor{linkblue}{rgb}{0.21,0.49,0.74}
\usepackage[breaklinks,colorlinks,allcolors=linkblue]{hyperref}
\def\systemname{mmTherm}

\title{\systemname: Absolute 3D Human Pose Estimation from \\Low-Resolution Thermal Images and mmWave Radar}
\author{%
Zhaoxin Chang\textsuperscript{1} \qquad
Badii Jouaber\textsuperscript{1} \qquad
Daqing Zhang\textsuperscript{1,2}\\[4pt]
{\small \textsuperscript{1}SAMOVAR, Telecom SudParis, Institut Polytechnique de Paris
\quad \textsuperscript{2}Peking University}%
}
\newcommand{\paperauthors}{Zhaoxin Chang, Badii Jouaber, Daqing Zhang}
\date{}
\hypersetup{
  pdftitle={mmTherm: Absolute 3D Human Pose Estimation from Low-Resolution Thermal Images and mmWave Radar},
  pdfauthor={\paperauthors},
  pdfsubject={Preprint},
  pdfkeywords={3D human pose estimation, thermal imaging, mmWave radar, multimodal learning}
}

\begin{document}
\maketitle
\begin{abstract}

Absolute 3D human pose estimation (HPE) aims to recover both a person's body pose and their global position in 3D space. Privacy-preserving millimeter-wave (mmWave) radar provides distance and direction measurement capabilities, but its sparse point clouds make HPE under-constrained. Low-resolution thermal imaging can provide complementary observations of body direction and structure. However, our evaluation shows that a strong feature-level fusion baseline loses much of its localization and HPE advantage when generalizing to unseen subjects, with radial localization becoming less accurate than a radar-only baseline. Motivated by this finding, we introduce \systemname{}, a lightweight framework that explicitly incorporates the complementary geometric properties of thermal imaging and mmWave radar. It combines body structure and joint directions estimated from thermal images with range information from radar to localize the body and reconstruct its 3D pose. Experiments on OctoNet show that \systemname{} reduces cross-user absolute joint error by 33\% and root localization error by 38\% relative to the strongest corresponding baselines, while requiring substantially fewer parameters and less computation at inference.

\end{abstract}

\section{Introduction}
\label{sec:introduction}

3D HPE plays an important role in understanding human activities and supports applications such as human-computer interaction, healthcare, and AR/VR. Traditional 3D HPE approaches rely on RGB cameras, which may degrade in low light and raise privacy concerns in settings such as smart homes. As an alternative, radio-frequency (RF) sensing has emerged as a promising tool for 3D HPE~\cite{adib_capturing_2015,zhao2018rfpose,ren_gopose_2022}, as RF signals reflected from human bodies carry information about human movement. One widely used RF sensing modality is mmWave radar~\cite{xue2021mmmesh,kong2022m3track,zhang_mupose_2026}. In practice, radar measurements are commonly processed into point clouds for 3D HPE~\cite{peng_expanding_nodate,wang_human_2023,yang_omnipc_2026}. 

\begin{figure*}[t]
  \centering
  \includegraphics[width=0.9\textwidth]{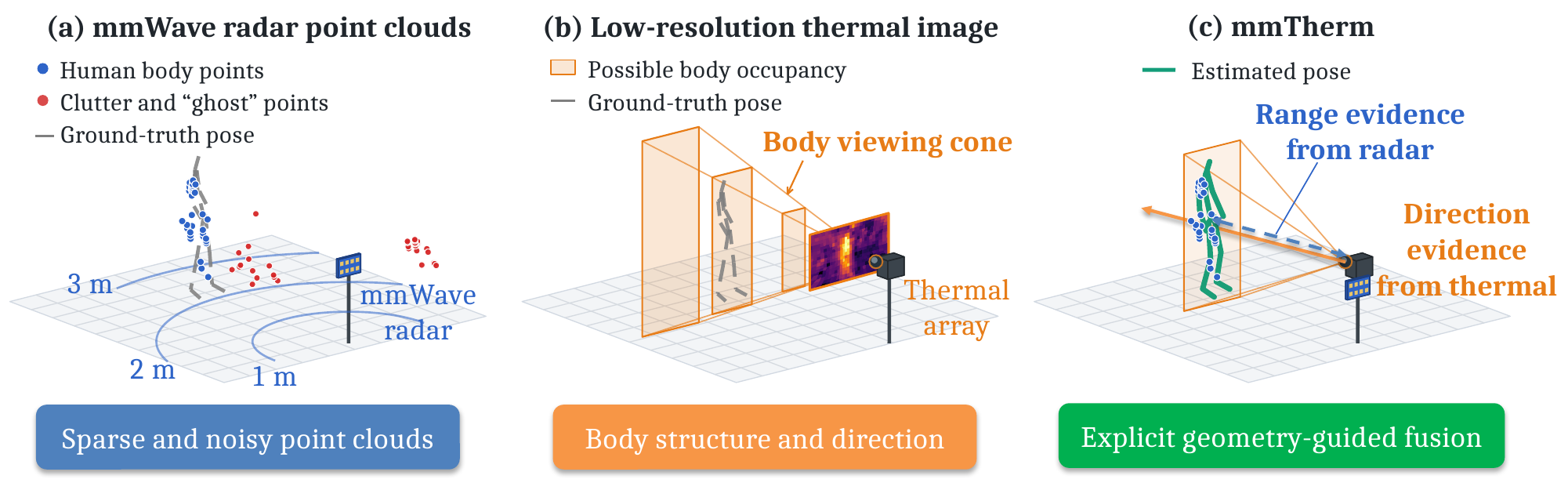}
  \caption{\textbf{Complementary geometry for absolute 3D HPE.} (a) Radar provides range information through sparse and noisy point clouds. (b) Thermal images provide coarse body structure and direction; orange planes illustrate depth ambiguity. (c) \systemname{} explicitly combines thermal direction and radar range evidence for pose reconstruction. }
  \label{fig:insight}
\end{figure*}

However, point clouds generated by mmWave radar are typically sparse and noisy. First, radar signal reflections from the target depend on posture, body orientation, and radar viewpoint, making some body parts have few or even no detectable returns~\cite{adib_capturing_2015,yang_spatial_2025,su2026mmwaveflow}. As a result, radar point clouds provide incomplete observations of the human body rather than dense measurements. Second, reflections from surrounding objects and multipath propagation can introduce background clutter and ``ghost'' points into point clouds~\cite{chen2023environment,kim2024multipath,zhang_pathclean_2026}. Consequently, radar point clouds neither provide complete coverage of the human body nor reliably indicate which returns correspond to individual body parts (\cref{fig:insight}a), making accurate 3D HPE challenging.

To address these challenges, existing works either improve the quality of mmWave point clouds~\cite{yang_mmdear_2025,su2026mmwaveflow,yang_omnipc_2026,yang_bridging_2026,zhang_pathclean_2026} or leverage deep learning methods to reconstruct human skeletons or meshes from sparse and noisy point clouds~\cite{xue_m4_2022,cao_occmesh_2025,fan2025mmdiff,fan2026m4human,liang2026wicompass,liang2026wave2body}. Although point-cloud enhancement and deep-learning methods can suppress noise and infer missing points, the physical limitations of radar visibility are not removed. Therefore, for body parts that cannot be directly observed, their positions must be inferred from limited measurements. Consequently, despite these advances, the ambiguity caused by incomplete and noisy radar observations remains challenging and makes 3D HPE an under-constrained problem.

Recent works have incorporated learned human pose priors into mmWave point cloud-based HPE~\cite{fan2025mmdiff,liang2026wave2body} to constrain pose reconstruction from incomplete radar observations. While such priors improve the plausibility of generated human poses, they do not provide independent observations to resolve missing or ambiguous radar returns. We therefore introduce thermal imaging as a complementary sensing modality that provides additional geometric observations. By capturing body heat, thermal images reveal coarse body structures and direction, providing spatial priors for body regions with sparse radar returns and helping distinguish human targets from background clutter (\cref{fig:insight}b). Specifically, we use a low-resolution (32$\times$24) thermal array for cost and privacy considerations. %

A natural way to incorporate these thermal observations into radar-based HPE is to fuse features from the two modalities. For example, MIRaSeg~\cite{tang2026miraseg} employs feature-level fusion to learn cross-modal representations, with a jointly trained pose estimation pathway. Indeed, our reproduction of its pose pathway (MIRaSeg-Pose) achieves strong in-domain absolute 3D HPE performance, outperforming radar-based baselines~\cite{fan2025mmdiff,liang2026wicompass,liang2026wave2body} in both body localization and relative pose estimation. Surprisingly, however, its localization advantage largely vanishes when generalizing to unseen subjects. In our cross-user evaluation, MIRaSeg-Pose achieves a root localization error of 156 mm, comparable to the 154 mm of the radar-based baseline Wave2Body~\cite{liang2026wave2body}. More notably, its radial localization error reaches 124 mm, more than twice that of Wave2Body (61 mm), despite achieving better directional accuracy. These findings suggest that implicit feature fusion, while effective within the training distribution, may learn dataset-specific relationships without reliably exploiting the distinct physical measurement capabilities of each modality. This motivates us to develop a fusion strategy that more explicitly accounts for the complementary geometric properties of the two modalities.

In this paper, we propose \systemname{}, a lightweight geometry-guided fusion framework for absolute 3D HPE using mmWave radar and low-resolution thermal imaging. As shown in~\cref{fig:insight}c, the framework explicitly incorporates thermal direction and body-region geometry into learned fusion with radar-derived position and range evidence. In the first stage, thermal geometry guides a learned correction to the radar-estimated root location, producing an estimate of the global body position. In the second stage, the fused root defines a local radar neighborhood and a spatial reference for pose reconstruction. Thermal features and calibrated geometry guide joint-level radar point association. Joint directions inferred from thermal images are then combined with range estimates from radar evidence to initialize the pose, which is subsequently refined by a lightweight spatiotemporal network.

We evaluate \systemname{} on OctoNet~\cite{yuan2025octonet}, a multimodal dataset with synchronized thermal and radar measurements and ground-truth 3D human poses. Under a cross-user split, \systemname{} reduces absolute joint error by 33\% and root localization error by 38\% relative to the best-performing corresponding baselines, while requiring substantially fewer parameters and less computation. Further experiments show that the proposed geometry-guided framework maintains its advantage in absolute pose accuracy under strong clutter and severe radar point sparsity. Joint-level analysis additionally shows improved reconstruction of body parts without nearby radar returns compared with radar-only estimation, supporting the value of complementary thermal observations under incomplete radar coverage. Our contributions can be summarized as follows:
\begin{itemize}
    \item We investigate a generalization challenge in radar--thermal fusion-based 3D HPE: a strong feature-level fusion baseline achieves accurate in-domain localization but loses its advantage on unseen subjects. This observation motivates explicitly exploiting the complementary geometric capabilities of thermal imaging and mmWave radar.
    \item We propose \systemname{}, a lightweight two-stage framework that combines explicit thermal geometry with radar-derived position and range for root localization, followed by root-conditioned pose reconstruction, without requiring point-level human or clutter annotations.
    \item We conduct extensive evaluations on OctoNet, demonstrating improvements in absolute 3D HPE under cross-user generalization and robustness to sparse and cluttered radar observations. %
\end{itemize}

\section{Related Work}
\label{sec:related}

\paragraph{3D HPE Using mmWave Radar.}
Early RF sensing systems such as RF-Pose demonstrate HPE capability of radio signals using visual supervision~\cite{zhao2018rfpose}. Subsequent systems, including mm-Pose, mmMesh, and MVDoppler-Pose, explore pose estimation and body reconstruction from mmWave radar observations~\cite{sengupta2020mmpose,xue2021mmmesh,choi2025mvdopplerpose,yang2023mmfi,ho2025rtpose}. However, sparse and noisy radar returns make pose reconstruction under-constrained~\cite{chen2022mmbody,fan2026m4human}. Existing efforts address these limitations through two broad directions. One improves radar point-cloud quality through signal processing or learned enhancement~\cite{yang_spatial_2025,yang_bridging_2026}. OmniPC jointly exploits range, Doppler, and angular information~\cite{yang_omnipc_2026}, while PathClean suppresses multipath during point-cloud generation~\cite{zhang_pathclean_2026}. Learning-based methods such as mmDEAR and mmWaveFlow further enhance or generate point clouds for downstream reconstruction~\cite{yang_mmdear_2025,su2026mmwaveflow}. The others develop deep learning methods for pose reconstruction, using human-body priors to improve inference plausibility or training coverage~\cite{xue_m4_2022,cao_occmesh_2025}. Specifically, mmDiff incorporates body structure and motion continuity into conditional diffusion~\cite{fan2025mmdiff}. WiCompass learns a reference pose space from motion-capture data to guide the collection of underrepresented poses~\cite{liang2026wicompass}. Wave2Body translates radar representations into a pretrained compositional body-token space~\cite{liang2026wave2body}. Although these approaches improve radar-based HPE performance, the positions of body parts without radar returns must still be inferred from incomplete observations. \systemname{} introduces complementary thermal measurements during inference, providing additional observations of coarse body structure and direction for absolute 3D HPE.

\paragraph{Thermal Sensing and Thermal--Radar Fusion.}
Low-resolution thermal arrays provide coarse spatial observations of the human body without conventional RGB imagery. TADAR investigates human detection and ranging~\cite{zhang2024tadar}, TAPOR reconstructs 3D hand poses~\cite{zhang2025tapor}, and other methods estimate body poses from ultra-low-resolution thermal images~\cite{kurihara2021boundary,arai2024thermalpose}. Since these arrays do not directly measure depth, recovering range or 3D pose requires additional modeling or learned priors. Radar--thermal fusion has also been explored with higher-resolution thermal cameras. UTM uses uncertainty-guided feature fusion for human detection under visual degradation~\cite{cai2023utm}; Zhao \etal{} associate infrared detections with radar tracks for nighttime pedestrian detection~\cite{zhao2023nighttime}; and RIDERS combines thermal images and radar measurements for dense metric depth estimation~\cite{li2024riders}. These thermal cameras retain much richer spatial detail than low-resolution arrays, placing them closer to conventional camera-based sensing. This does not match our design goal of low-cost sensing with deliberately limited visual detail, motivated by the resolution--privacy--cost trade-off of thermal sensors~\cite{zhang2024tadar,rintahomi2021howlow}. More closely related, MIRaSeg fuses mmWave radar point clouds and low-resolution infrared images for human semantic segmentation, with a jointly trained pose pathway~\cite{tang2026miraseg}. \systemname{} is designed for absolute 3D HPE with a $32\times24$ thermal array, explicitly using calibrated thermal directions and body regions to guide radar point association and pose reconstruction.

\section{Method}
\label{sec:method}

\subsection{Overview}

\paragraph{Problem Formulation.}
\systemname{} estimates an absolute 20-joint pose from synchronized low-resolution thermal images and mmWave radar point clouds. The co-located sensors are calibrated in a common radar coordinate system, in which we denote the predicted pose as $\widehat{\mathcal P}=\{\widehat{\mathbf p}_j\}_{j=1}^{20}$. We use the pelvis as the root joint throughout this work.

\begin{figure*}[t]
  \centering
  \includegraphics[width=0.9\textwidth]{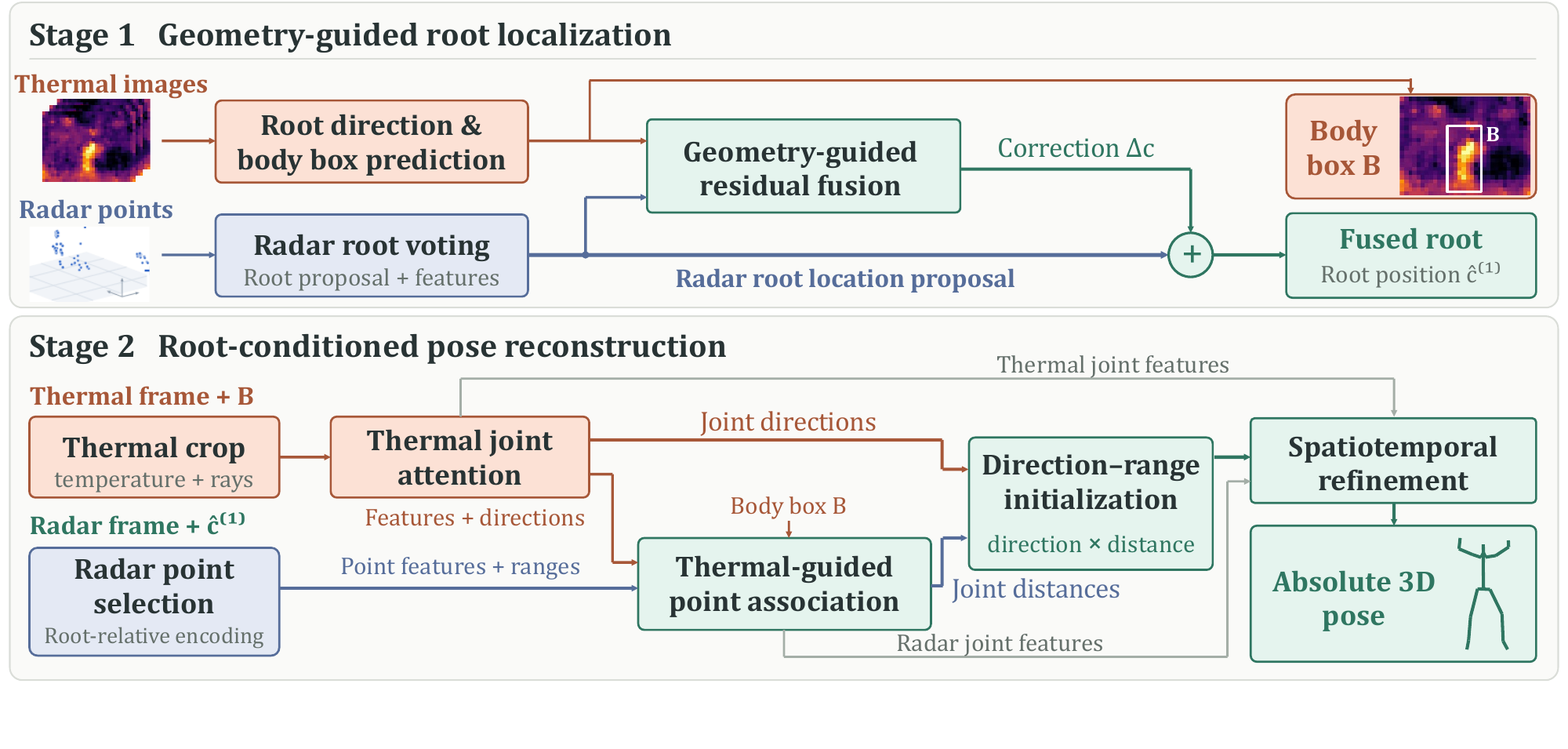}
  \caption{\textbf{Overview of \systemname{}.} Stage~1 combines thermal geometry with radar observations to localize the root. Using the predicted root and body box, Stage~2 estimates joint directions from thermal images and joint ranges through thermal-guided radar attention. These estimates form an initial 3D pose, which is then refined by a spatiotemporal network.}
  \label{fig:architecture}
\end{figure*}

\paragraph{\systemname{}.}
Absolute 3D HPE requires both global body localization and articulation recovery. \systemname{} achieves these through two stages of geometry-guided fusion (\figref{fig:architecture}). Stage~1 combines thermal direction and body-region geometry with radar position and range to localize the root joint. Stage~2 selects radar points around the fused root and uses thermal features and geometry to associate them with individual joints. It then combines thermal-derived joint directions with radar-supported range measurements to initialize the pose, followed by a spatiotemporal refinement module.

\subsection{Geometry-Guided Root Joint Localization}
\label{sec:localization}

Stage~1 first processes the two modalities through separate branches: the thermal branch estimates the root direction and body bounding box, while the radar branch predicts a 3D root position proposal from the point cloud. A fusion head then combines their outputs, using thermal geometry to guide a learned spatial correction to the radar proposal.

\paragraph{Thermal-Based Direction and Body Region Estimation.}
The thermal branch estimates the root direction and a coarse body region to guide radar-based localization and subsequent pose reconstruction. A convolutional encoder takes three consecutive normalized thermal frames and their per-frame validity flags, broadcast as image channels. It predicts a root heatmap and body bounding box $B$, supervised by the projected ground-truth root and the bounding box of the projected skeleton. Soft-argmax yields the root location in the image. Using calibrated viewing rays, we convert this location into a unit direction $\widehat{\mathbf b}_T$ and the box into an angular center with horizontal and vertical spans. These outputs describe the person's direction and angular coverage without assigning depth.

\paragraph{Radar-Based Root Voting.}
Due to the sparsity and ambiguity, radar point clouds cannot be directly used to locate the root joint. This radar branch estimates the 3D position of the root joint by aggregating learned votes from individual points. It takes valid points from three temporal frames and encodes each point $\mathbf r_i$ using its 3D coordinates, distance, Doppler velocity, and relative time, together with global mean- and max-pooled context. For each point, the branch predicts an offset $\boldsymbol\delta_i$ toward a candidate root location and a contribution score $a_i$. These predictions yield the root position proposal:
\begin{equation}
  \alpha_i=\operatorname{softmax}_i(a_i),
  \qquad
  \widehat{\mathbf c}_R
  =\sum_i\alpha_i(\mathbf r_i+\boldsymbol\delta_i).
  \label{eq:root_voting}
\end{equation}
The weights are normalized over valid points. Both the aggregated proposal and individual votes are supervised by the ground-truth root position, without requiring point-level human or clutter annotations.

\paragraph{Geometry-Guided Fusion.}
The fusion head combines the radar branch-proposed position and range with thermal direction and body-region geometry to refine root localization. To represent their geometric relationship explicitly, we construct a descriptor $\mathbf g$ containing $\widehat{\mathbf c}_R$, $\|\widehat{\mathbf c}_R\|_2$, pooled radar features, the directions $\widehat{\mathbf b}_T$ and $\widehat{\mathbf b}_R=\widehat{\mathbf c}_R/\|\widehat{\mathbf c}_R\|_2$, their dot and cross products, and the angular spans of the thermal box. The dot and cross products encode directional agreement and discrepancy between the two estimates.

To describe observation quality and availability, $\mathbf g$ also includes the maximum voting weight, normalized voting entropy, valid-point fraction, thermal heatmap peak, and thermal-frame validity flags. The unweighted centroid, log point count, and spatial spread additionally summarize the valid point cloud independently of learned voting weights. From this descriptor, an MLP predicts a spatial correction to the radar root location proposal:
\begin{equation}
  \Delta\mathbf c=f_{\theta}(\mathbf g),
  \qquad
  \widehat{\mathbf c}^{(1)}
  =\widehat{\mathbf c}_R+\Delta\mathbf c.
  \label{eq:root_fusion}
\end{equation}
The correction can adjust all three coordinates, so radar range is not fixed. The resulting fusion-based root position provides a spatial reference for the next stage.

\subsection{Root-Conditioned Pose Reconstruction}
\label{sec:association}

Stage~2 reconstructs the full 3D pose from a thermal crop defined by $B$ and radar points selected around the fused root location $\widehat{\mathbf c}^{(1)}$. Since low-resolution thermal pixels and radar point clouds do not directly identify skeletal joints, the model learns joint-specific associations to estimate joint directions from thermal observations and joint ranges from radar evidence. These estimates initialize a 3D pose, which is then refined across joints and time.

\paragraph{Local Thermal and Radar Representations.}
To focus pose reconstruction on the person, we use the Stage~1 predictions to select the thermal body region and nearby radar points while preserving their geometric information. For thermal images, we expand $B$ by one pixel on each side to include a small margin around the predicted body region and resample the crop to $32\times32$. The thermal input has four channels: temperature standardized within the crop and the three components of the calibrated viewing direction. Temperature and direction are resampled together to preserve their spatial correspondence.

For each thermal frame, we process its paired radar point cloud and retain valid points within 1.5~m of $\widehat{\mathbf c}^{(1)}$. Each retained point is represented by seven features: its 3D coordinates relative to the fused root, its distance to that root, and the three components of its unit direction from the radar origin. This encoding describes the point's location relative to the person while preserving its viewing direction. We separately retain each point's range from the radar origin for joint-range estimation.

\paragraph{Thermal-Estimated Joint Directions.}
We estimate each joint's viewing direction from the thermal crop to guide radar association and pose initialization. A thermal encoder produces an $8\times8$ grid of patch features. Twenty learned joint queries attend to these patches to obtain joint-specific thermal features. For each joint $j$, we average the attention weights across heads and use them to aggregate the calibrated patch directions. Normalizing the resulting vector yields the estimated unit direction $\widehat{\mathbf d}_j$.

\paragraph{Thermal-Guided Radar Attention.}
We use thermal features and geometry to weight radar points according to their relevance to each joint. The radar attention query for joint $j$ is formed by adding a linear projection of its thermal feature to its learned joint query. The attention score for radar point $i$ combines feature compatibility with directional alignment and a body-region penalty:
\begin{equation}
  \ell_{ji}=s_{ji}+\beta\cos\theta_{ji}-\gamma P_i.
  \label{eq:attention}
\end{equation}
Here, $s_{ji}$ is the scaled query--key score, and $\theta_{ji}$ is the angle between the estimated joint direction $\widehat{\mathbf d}_j$ and the point's direction from the radar origin. The penalty $P_i$ discourages attention to points whose directions fall outside the angular region defined by the thermal body box. The learned coefficients $\beta$ and $\gamma$ are constrained to be nonnegative through softplus and are shared across attention heads. The penalty is zero inside the angular box. Outside it, we normalize the azimuth and elevation excesses by twice the corresponding median angular spacing between neighboring pixel rays. The penalty is half the sum of the squared normalized excesses, capped at six. Softmax normalization then produces joint-specific attention weights for aggregating radar features and estimating joint ranges.

\paragraph{Direction--Range Initialization.}
We combine joint directions estimated from thermal images with ranges inferred from radar points to construct an initial absolute 3D pose. We estimate each joint's range $\widehat{\rho}_j$ as an attention-weighted sum of point ranges and a fallback range. The fallback is supplied by a learned root token that participates in the radar attention softmax without either geometric bias in \cref{eq:attention}. Its range is $\|\widehat{\mathbf c}^{(1)}\|_2$ at inference and $\|\widetilde{\mathbf c}\|_2$ during training (\secref{sec:learning}). Combining range with thermal direction initializes every joint, including the root:
\begin{equation}
  \widehat{\mathbf p}^{\mathrm{init}}_j
  =\widehat{\rho}_j\widehat{\mathbf d}_j.
  \label{eq:init}
\end{equation}
Direction and range share the radar origin as their reference.

\paragraph{Spatiotemporal Refinement.}
We refine this approximate initialization across joints and time to account for coarse thermal resolution and sparse radar reflections that do not directly locate skeletal joints. Each joint receives the attention-weighted aggregation of retained point directions, excluding the root token. This radar direction falls back to the direction of $\widehat{\mathbf c}_R$ only when no point is retained.

Joint tokens combine thermal and radar attention features, the encoded radar direction and fallback indicator, estimated range, and the dot product between thermal and radar directions. They also contain root-relative initialized coordinates, $\widehat{\mathbf p}^{\mathrm{init}}_j-\widehat{\mathbf p}^{\mathrm{init}}_{\mathrm{root}}$. With a body token, they pass through a spatial--temporal--spatial Transformer (depths 1--2--1) that exchanges information across joints and three frames, uses actual observation-time differences, and predicts coordinate corrections:
\begin{equation}
  \widehat{\mathbf p}_j
  =\widehat{\mathbf p}^{\mathrm{init}}_j+\Delta\mathbf p_j.
  \label{eq:pose_refinement}
\end{equation}
Although token coordinates are root-relative, corrections are added to the absolute initialization. The final root $\widehat{\mathbf p}_{\mathrm{root}}$ is refined with the other joints and may differ from Stage~1.

\subsection{Training Objectives and Optimization}
\label{sec:refinement}
\label{sec:learning}

\paragraph{Staged Optimization.}
We first train the thermal branch with projected root and box supervision and the radar branch with 3D root supervision, including unweighted per-point vote losses. With both branches frozen, we train the fusion MLP using root-position error, Smooth L1 regression, and a correction-magnitude penalty weighted by $0.01$. Its final layer is initialized near zero. Stage~2 is then trained and evaluated using predicted roots and boxes from the frozen Stage~1 model.

\paragraph{Pose Objectives.}
We supervise root-relative pose and absolute root location separately, together with the intermediate joint directions and ranges. The four objectives have equal weights:
\begin{equation}
  \mathcal L
  =\mathcal L_{\mathrm{rel}}
  +\mathcal L_{\mathrm{root}}
  +\mathcal L_{\mathrm{dir}}
  +\mathcal L_{\mathrm{range}}.
  \label{eq:training_loss}
\end{equation}
The losses $\mathcal L_{\mathrm{rel}}$ and $\mathcal L_{\mathrm{root}}$ supervise root-aligned joint positions and the final root position, respectively, using Smooth L1 with a transition scale of 0.06~m. The direction loss measures cosine agreement, while the range loss uses Smooth L1 with a transition scale of 0.08~m. All targets are derived from the ground-truth 3D skeleton.

\paragraph{Training Augmentations.}
To expose Stage~1 to sparse radar observations, half of the radar-branch and fusion-head training samples undergo radar point removal, with retention ratios sampled uniformly from $[0.1,1]$. For Stage~2, we perturb the conditioning roots as $\widetilde{\mathbf c}_t=\widehat{\mathbf c}^{(1)}_t+\boldsymbol\epsilon$, where one zero-mean Gaussian offset ($\sigma=0.05$~m per axis) is shared across all frames of a training clip. The perturbed roots affect only root-relative point encoding and the fallback range; cached radar points remain selected around the unperturbed roots. This discourages direct copying of the conditioning root into the final prediction. Both augmentations are disabled during validation and testing.
\begin{figure*}[t]
  \centering
  \includegraphics[width=0.85\textwidth]{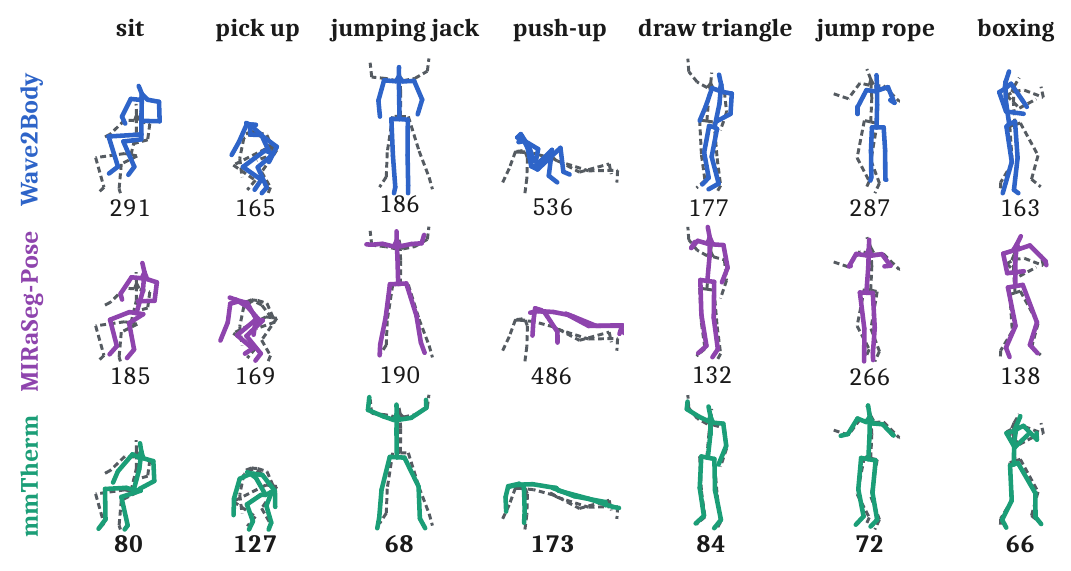}
  \caption{\textbf{Qualitative HPE results on unseen users.} Each column uses the same frame across methods. Dashed skeletons are ground-truth poses. Values indicate absolute MPJPE in mm. The best result in each column is bold.}
  \label{fig:qualitative}
\end{figure*}

\section{Experiments}
\label{sec:experiments}

\subsection{Experimental Setup}
\label{sec:protocol}

\paragraph{Dataset.}
We evaluate \systemname{} on OctoNet~\cite{yuan2025octonet}, which provides mmWave point clouds and $32\times24$ thermal images from five nodes, with OptiTrack 3D pose ground truth. Recordings capture one person repeating an activity about ten times (median 67~s). Each prediction uses one radar--thermal pair, with training samples pooled across nodes.

\paragraph{Evaluation Protocols.}
For \emph{in-domain} evaluation, recordings within each activity are split into training, validation, and test sets at a 7/1/2 ratio (254,174 test samples). \emph{Cross-user} evaluation trains on users~1--7, validates on user~8, and tests on users~9--10 (233,518 samples). All node views and overlapping windows of a recording remain in the same split.
We further evaluate \emph{unseen users and activities} using Gym and Freestyle recordings from 29 additional people absent from the main splits (456,802 samples). These activities are excluded from training and checkpoint selection. 

\paragraph{Metrics.}
We report errors in millimeters. Absolute (ABS) mean per joint position error (MPJPE) averages joint errors in the coordinate system. Root error measures pelvis-localization error. Root-aligned (RA) MPJPE subtracts each pose's root without rotation or scale alignment. Both MPJPE metrics include all 20 joints. We also compare lateral root error, perpendicular to the true line of sight, and range error.

\paragraph{Comparisons.}
We compare \systemname{} with mmDiff~\cite{fan2025mmdiff}, WiCompass~\cite{liang2026wicompass}, Wave2Body~\cite{liang2026wave2body}, and MIRaSeg-Pose~\cite{tang2026miraseg} under identical splits.  The reproduced mmDiff pathway predicts root-relative poses (RA-MPJPE) only. MIRaSeg-Pose is our adaptation of MIRaSeg’s pose pathway to OctoNet’s absolute 20-joint skeleton, trained with pose supervision without segmentation labels.

\paragraph{Implementation Details.}
We train on an NVIDIA RTX PRO 6000 Blackwell Max-Q GPU. Stage~2 uses 128-dimensional tokens and four attention heads. We train Stage~2 and its ablations with AdamW (learning rate $10^{-4}$, weight decay 0.01), cosine decay over at most 60 epochs, and batches of 24 clips in FP32. We select the checkpoint with the lowest ABS MPJPE on the validation set.

\subsection{Main Results}
\label{sec:mainresults}

\Cref{tab:main} shows that \method{} outperforms all radar-only baselines on the available pose metrics under both protocols. Compared with MIRaSeg-Pose, in-domain absolute and root-aligned errors are comparable, while root localization improves. The advantage grows on unseen users, where \method{} achieves lower errors on all three metrics, reducing absolute MPJPE and root error by 33\% and 39\%, respectively. The qualitative results in \cref{fig:qualitative} illustrate the benefits of incorporating thermal observations: both radar--thermal methods recover body postures more accurately than radar-only Wave2Body. Compared with MIRaSeg-Pose, \method{} further improves both absolute body placement and limb articulation, as illustrated by the jumping-jack and jump-rope examples, producing skeletons that more closely match the references.

To test the combined shift in users and activities, we further evaluate frozen in-domain models on Gym and Freestyle. \method{} achieves an absolute MPJPE of 193.2~mm and a root error of 138.0~mm, compared with 225.3~mm and 177.9~mm for MIRaSeg-Pose.

\begin{table}[t]
  \centering
  \small
  \setlength{\tabcolsep}{2.5pt}
  \caption{\textbf{3D HPE errors on OctoNet}
  (mm, lower is better). ABS: absolute MPJPE; Root: root position error;
  RA: root-aligned MPJPE.}
  \begin{tabular}{@{}lcccccc@{}}
    \toprule
    & \multicolumn{3}{c}{In-domain} & \multicolumn{3}{c}{Cross-user} \\
    \cmidrule(lr){2-4}\cmidrule(lr){5-7}
    Method & ABS & Root & RA & ABS & Root & RA \\
    \midrule
    mmDiff~\cite{fan2025mmdiff}
      & -- & -- & 126.5 & -- & -- & 135.6 \\
    WiCompass~\cite{liang2026wicompass}
      & 173.1 & 149.6 & 103.2 & 265.5 & 249.3 & 148.5 \\
    Wave2Body~\cite{liang2026wave2body}
      & 129.4 & 96.8 & 91.6 & 187.1 & 153.9 & 129.3 \\
    MIRaSeg-Pose~\cite{tang2026miraseg}
      & 80.4 & 62.7 & \textbf{63.7} & 178.1 & 156.4 & 111.1 \\
    \midrule
    \method{} (ours)
      & \textbf{80.2} & \textbf{57.3} & 66.2
      & \textbf{118.7} & \textbf{95.1} & \textbf{98.8} \\
    \bottomrule
  \end{tabular}
  \label{tab:main}
\end{table}

\subsection{Robustness Evaluation}
\label{sec:clutter}

We next test whether the cross-user gains persist under ambiguous or incomplete radar observations, using strong clutter, point removal, and a missing-return diagnostic.

\begin{figure}[b]
  \centering
  \includegraphics[width=1\linewidth]{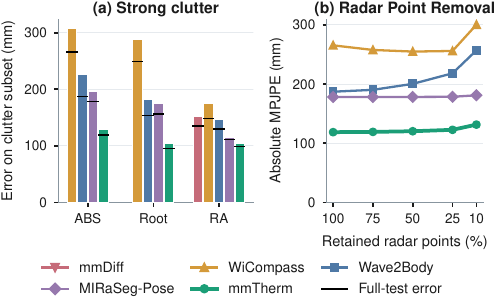}
  \caption{\textbf{Robustness evaluation.} (a) The top 20\% of recordings by the clutter ranking; black ticks show full-test errors. (b) Absolute MPJPE after radar point removal.}
  \label{fig:robust}
\end{figure}

\paragraph{Strong Clutter.}
We select the 20\% of cross-user recordings with the highest mean count of clustered radar points at least 1~m from the reference skeleton (0.35~m neighborhood radius; at least three points per cluster). Ground-truth poses serve only for offline selection. \method{} achieves the lowest root-aligned and absolute errors on this subset (\cref{fig:robust}a). The proxy does not control cluster origins or other recording differences.

\paragraph{Radar Point Removal.}
We evaluate frozen models after randomly removing radar points before preprocessing, keeping thermal inputs unchanged (\cref{fig:robust}b). At 10\% retention, radar-only baselines show substantial increases in absolute MPJPE, while \method{} exhibits a more moderate increase. MIRaSeg-Pose is the least sensitive to point removal, but \method{} maintains lower absolute MPJPE at every retention level. This limited sensitivity may partly reflect weaker reliance on radar-derived range information in MIRaSeg-Pose's feature-level fusion, consistent with its larger root and absolute errors on unseen users. %

\begin{figure}[b]
  \centering
  \includegraphics[width=0.85\linewidth]{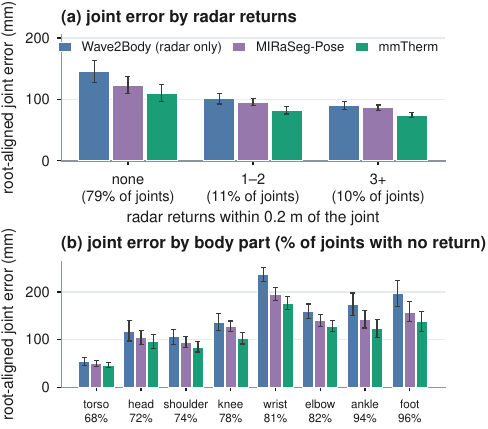}
  \caption{\textbf{Missing-return diagnostic}. (a) Root-aligned error by radar point count within 0.2~m of reference joints. (b) Root-aligned error by body part. Error bars: 95\% recording-bootstrap intervals.}
  \label{fig:jointsparsity}
\end{figure}

\paragraph{Missing Radar Returns.}
To examine pose reconstruction when local radar returns are missing, we count radar points within 0.2~m of each ground-truth joint location. Across the sampled frames, 79\% of non-root joint locations have no points within this radius. Both radar--thermal methods achieve their largest root-aligned improvements over Wave2Body for these joints (\cref{fig:jointsparsity}a). The largest gains occur at the wrists, feet, and ankles (\cref{fig:jointsparsity}b). %

\subsection{Ablation Studies and Analysis}
\label{sec:where}

To understand the performance gains, we analyze root localization and pose reconstruction to understand the contributions of the two stages.

\paragraph{Stage~1: Root Localization and Fusion.}
We first examine the final root predictions of the complete models in \cref{tab:rootsplit}. MIRaSeg-Pose achieves good lateral accuracy, but its cross-user range error is approximately twice that of Wave2Body. \method{} achieves the lowest range error under both protocols and the lowest lateral error on unseen users. These results suggest that effective radar--thermal fusion requires combining improved directional localization with reliable range estimation, particularly for unseen users. We therefore compare Stage~1 fusion strategies to examine the contribution of explicit thermal geometry.

To examine the role of root fusion, \cref{tab:stage1} compares the radar-only root proposal with feature fusion and our geometry-guided fusion. \emph{Feature fusion} replaces explicit thermal geometry with encoder features and a node indicator while retaining the radar residual. Both fusion variants improve root localization over the radar-only baseline. Compared with feature fusion, our geometry-guided fusion further reduces lateral error while maintaining comparable range accuracy, lowering the 3D root error by 10.8~mm. These results support explicitly incorporating thermal geometry into root fusion.

\begin{table}[t]
  \centering
  \small
  \setlength{\tabcolsep}{3.4pt}
  \caption{\textbf{Final root localization errors across methods on 20k sampled test frames} (mm, lower is better).}
  \begin{tabular}{lcccccc}
    \toprule
    & \multicolumn{3}{c}{In-domain} & \multicolumn{3}{c}{Cross-user} \\
    \cmidrule(lr){2-4}\cmidrule(lr){5-7}
    Method & Lat. & Range & 3D & Lat. & Range & 3D \\
    \midrule
    Wave2Body & 75.6 & 47.2 & 96.1 & 131.7 & 60.6 & 154.0 \\
    MIRaSeg-Pose & \textbf{32.9} & 47.3 & 62.8 & 78.5 & 123.7 & 156.2 \\
    \method{} & 36.4 & \textbf{36.9} & \textbf{57.0} & \textbf{67.0} & \textbf{55.7} & \textbf{95.0} \\
    \bottomrule
  \end{tabular}
  \label{tab:rootsplit}
\end{table}

\begin{table}[t]
  \centering
  \small
  \setlength{\tabcolsep}{5pt}
  \caption{\textbf{Comparison of Stage~1 root fusion strategies on unseen users} (mm, lower is better).}
  \begin{tabular}{lccc}
    \toprule
    Stage~1 variant & Lateral & Range & 3D \\
    \midrule
    Radar only & 123.0 & 59.9 & 145.1 \\
    Feature fusion & 84.2 & 56.9 & 110.7 \\
    Full fusion & \textbf{71.3} & \textbf{56.4} & \textbf{99.9} \\
    \bottomrule
  \end{tabular}
  \label{tab:stage1}
\end{table}

\paragraph{Stage~2: Pose Reconstruction.}
With Stage~1 conditioning fixed, we examine the cross-user results in \cref{tab:ablation}. Removing Stage~2 radar points increases absolute MPJPE from 118.7 to 143.6~mm, while removing thermal temperatures increases it to 140.4~mm. Both ablations also worsen root localization and root-aligned pose accuracy. Jointly removing direction--range initialization and its auxiliary direction and range losses increases absolute MPJPE to 127.8~mm, with both root and root-aligned errors also increasing. These results support combining complementary radar and thermal observations with geometrically supervised initialization for pose reconstruction.

\begin{table}[t]
  \centering
  \small
  \setlength{\tabcolsep}{3.2pt}
  \caption{\textbf{Stage~2 ablations} (mm, lower is better). Removing temperatures preserves the thermal box and calibrated rays.}
  \begin{tabular}{lcccccc}
    \toprule
    & \multicolumn{3}{c}{In-domain} & \multicolumn{3}{c}{Cross-user} \\
    \cmidrule(lr){2-4}\cmidrule(lr){5-7}
    Stage~2 variant & ABS & Root & RA & ABS & Root & RA \\
    \midrule
    Full model
      & \textbf{80.2} & \textbf{57.3} & 66.2
      & \textbf{118.7} & \textbf{95.1} & \textbf{98.8} \\
    w/o S2 radar points
      & 100.3 & 74.1 & 87.1
      & 143.6 & 104.1 & 126.3 \\
    w/o temperatures
      & 99.2 & 64.6 & 84.6
      & 140.4 & 107.8 & 116.6 \\
    w/o geometric init.
      & 83.1 & 62.0 & \textbf{65.5}
      & 127.8 & 98.2 & 101.0 \\
    \bottomrule
  \end{tabular}
  \label{tab:ablation}
\end{table}

\subsection{Model Cost}
\label{sec:cost}

We compare model size, computational cost, and inference throughput in \cref{tab:cost}. Parameter counts include all components used at inference, and FLOPs are reported per pose as $2\times$MACs. Throughput is measured in FP32 with a batch size of 256. \method{} uses the fewest parameters and FLOPs and achieves the highest throughput among the evaluated methods. Compared with MIRaSeg-Pose, it reduces the parameter count by a factor of $8.5$ and the FLOPs by a factor of $11$, while achieving $2.4\times$ the throughput. These results demonstrate improved cross-user accuracy with a smaller model and lower inference computation.

\begin{table}[htbp]
  \centering
  \small
  \setlength{\tabcolsep}{4pt}
  \caption{\textbf{Model cost and throughput.} }
  \begin{tabular}{lccc}
    \toprule
    Method & Params (M) $\downarrow$ & GFLOPs $\downarrow$ & Throughput $\uparrow$ \\
    \midrule
    mmDiff & 32.09 & 14.09 & 873 \\
    WiCompass & 26.91 & 13.98 & 1,321 \\
    Wave2Body & 16.91 & 0.66 & 13,006 \\
    MIRaSeg-Pose & 11.64 & 3.63 & 8,988 \\
    \method{} & \textbf{1.37} & \textbf{0.33} & \textbf{21,397} \\
    \bottomrule
  \end{tabular}
  \label{tab:cost}
\end{table}
\section{Conclusion}
\label{sec:conclusion}

We presented \method{}, a geometry-guided framework for absolute 3D HPE from mmWave radar and a low-resolution thermal array. It uses thermal direction and body structure to guide the interpretation of radar-derived position and range evidence. Root localization establishes a spatial reference for joint-level reconstruction and spatiotemporal refinement. Experiments on OctoNet demonstrate improved absolute pose accuracy and root localization on unseen users, while maintaining an advantage under clutter and sparse radar observations with fewer parameters and lower inference cost.

\section*{Acknowledgments}
We gratefully acknowledge NVIDIA Corporation for providing the RTX PRO 6000 Blackwell Max-Q GPU through the NVIDIA Academic Grant Program.
{\small
\bibliographystyle{ieeenat_fullname}
\bibliography{references}

@inproceedings{yuan2025octonet,
  title = {{OctoNet}: A Large-Scale Multi-Modal Dataset for Human Activity Understanding Grounded in Motion-Captured 3D Pose Labels},
  author = {Yuan, Dongsheng and Zhang, Xie and Hou, Weiying and Lyu, Sheng and Yu, Yuemin and Yu, Luca Jiang-Tao and Li, Chengxiao and Wu, Chenshu},
  booktitle = {Advances in Neural Information Processing Systems},
  volume = {38},
  year = {2025},
  url = {https://proceedings.neurips.cc/paper_files/paper/2025/hash/14950adea25005f89545688fe97fe5ea-Abstract-Datasets_and_Benchmarks_Track.html}
}

@inproceedings{tang2026miraseg,
  title = {{MIRaSeg}: Exploring mmWave Radar and Low Resolution Infrared Sensor Fusion for Robust Human Semantic Segmentation},
  author = {Tang, Xiangjie and Shi, Ruili and Wang, Shuai and Zhang, Zeyu and Wang, Bin and Li, Zhiqiang and Wang, Shuai},
  booktitle = {Pattern Recognition and Computer Vision -- PRCV 2025},
  series = {Lecture Notes in Computer Science},
  volume = {16289},
  pages = {17--32},
  publisher = {Springer},
  year = {2026},
  doi = {10.1007/978-981-95-5764-6_2}
}

@inproceedings{fan2025mmdiff,
  title = {Diffusion Model is a Good Pose Estimator from 3D {RF-Vision}},
  author = {Fan, Junqiao and Yang, Jianfei and Xu, Yuecong and Xie, Lihua},
  booktitle = {Computer Vision -- ECCV 2024},
  series = {Lecture Notes in Computer Science},
  volume = {15074},
  pages = {1--18},
  publisher = {Springer},
  year = {2025},
  doi = {10.1007/978-3-031-72640-8_1}
}

@inproceedings{liang2026wicompass,
author = {Liang, Bo and Gong, Chen and Wang, Haobo and Liu, Qirui and Zhou, Rungui and Shao, Fengzhi and Wang, Yubo and Gao, Wei and Zhou, Kaichen and Cui, Guolong and Xu, Chenren},
title = {WiCompass: Oracle-driven Data Scaling for mmWave Human Pose Estimation},
year = {2026},
isbn = {9798400725050},
publisher = {Association for Computing Machinery},
address = {New York, NY, USA},
url = {https://doi.org/10.1145/3795866.3796684},
doi = {10.1145/3795866.3796684},
booktitle = {Proceedings of the 32nd Annual International Conference on Mobile Computing and Networking},
pages = {194–210},
numpages = {17},
location = {Austin, TX, USA},
series = {MobiCom '26}
}

@article{liang2026wave2body,
  title = {{Wave2Body}: Rethinking mmWave Human Pose Estimation as Radar-to-Body Token Translation},
  author = {Liang, Bo and Gong, Chen and Gao, Wei and Xu, Chenren},
  journal = {arXiv preprint arXiv:2607.18875},
  year = {2026},
  doi = {10.48550/arXiv.2607.18875},
  url = {https://arxiv.org/abs/2607.18875}
}

@inproceedings{zhao2018rfpose,
  title = {Through-Wall Human Pose Estimation Using Radio Signals},
  author = {Zhao, Mingmin and Li, Tianhong and Abu Alsheikh, Mohammad and Tian, Yonglong and Zhao, Hang and Torralba, Antonio and Katabi, Dina},
  booktitle = {Proceedings of the IEEE Conference on Computer Vision and Pattern Recognition},
  pages = {7356--7365},
  year = {2018},
  url = {https://openaccess.thecvf.com/content_cvpr_2018/papers/Zhao_Through-Wall_Human_Pose_CVPR_2018_paper.pdf}
}

@article{sengupta2020mmpose,
  title = {{mm-Pose}: Real-Time Human Skeletal Posture Estimation Using {mmWave} Radars and {CNNs}},
  author = {Sengupta, Arindam and Jin, Feng and Zhang, Renyuan and Cao, Siyang},
  journal = {IEEE Sensors Journal},
  volume = {20},
  number = {17},
  pages = {10032--10044},
  year = {2020},
  doi = {10.1109/JSEN.2020.2991741}
}

@inproceedings{xue2021mmmesh,
  title = {{mmMesh}: Towards 3D Real-Time Dynamic Human Mesh Construction Using Millimeter-Wave},
  author = {Xue, Hongfei and Ju, Yan and Miao, Chenglin and Wang, Yijiang and Wang, Shiyang and Zhang, Aidong and Su, Lu},
  booktitle = {Proceedings of the 19th Annual International Conference on Mobile Systems, Applications, and Services},
  pages = {269--282},
  year = {2021},
  doi = {10.1145/3458864.3467679}
}

@inproceedings{chen2022mmbody,
  title = {{mmBody} Benchmark: 3D Body Reconstruction Dataset and Analysis for Millimeter Wave Radar},
  author = {Chen, Anjun and Wang, Xiangyu and Zhu, Shaohao and Li, Yanxu and Chen, Jiming and Ye, Qi},
  booktitle = {Proceedings of the 30th ACM International Conference on Multimedia},
  pages = {3501--3510},
  year = {2022},
  url = {https://chen3110.github.io/mmbody/}
}

@inproceedings{su2026mmwaveflow,
  title = {{mmWaveFlow}: Unified Enhancement and Generation of {mmWave} Human Point Clouds},
  author = {Su, Chang and Jin, Beihong and Shi, Qiwen and Wang, Zhi},
  booktitle = {Proceedings of the IEEE/CVF Conference on Computer Vision and Pattern Recognition},
  pages = {31366--31376},
  year = {2026},
  url = {https://openaccess.thecvf.com/content/CVPR2026/html/Su_mmWaveFlow_Unified_Enhancement_and_Generation_of_mmWave_Human_Point_Clouds_CVPR_2026_paper.html}
}

@inproceedings{fan2026m4human,
  title = {{M4Human}: A Large-Scale Multimodal {mmWave} Radar Benchmark for Human Mesh Reconstruction},
  author = {Fan, Junqiao and Zhou, Yunjiao and Yang, Yizhuo and Cui, Xinyuan and Zhang, Jiarui and Xie, Lihua and Yang, Jianfei and Lu, Chris Xiaoxuan and Ding, Fangqiang},
  booktitle = {Proceedings of the IEEE/CVF Conference on Computer Vision and Pattern Recognition},
  pages = {42836--42846},
  year = {2026},
  url = {https://openaccess.thecvf.com/content/CVPR2026/html/Fan_M4Human_A_Large-Scale_Multimodal_mmWave_Radar_Benchmark_for_Human_Mesh_CVPR_2026_paper.html}
}

@inproceedings{zhang2024tadar,
  title = {{TADAR}: Thermal Array-based Detection and Ranging for Privacy-Preserving Human Sensing},
  author = {Zhang, Xie and Wu, Chenshu},
  booktitle = {Proceedings of the Twenty-fifth International Symposium on Theory, Algorithmic Foundations, and Protocol Design for Mobile Networks and Mobile Computing},
  year = {2024},
  doi = {10.1145/3641512.3686357}
}

@article{zhang2025tapor,
  title = {{TAPOR}: 3D Hand Pose Reconstruction with Fully Passive Thermal Sensing for Around-Device Interactions},
  author = {Zhang, Xie and Li, Chengxiao and Wu, Chenshu},
  journal = {Proceedings of the ACM on Interactive, Mobile, Wearable and Ubiquitous Technologies},
  volume = {9},
  number = {2},
  articleno = {63},
  numpages = {25},
  year = {2025},
  doi = {10.1145/3729499}
}

@inproceedings{arai2024thermalpose,
  title = {3D Human Pose Estimation Using Ultra-low Resolution Thermal Images},
  author = {Arai, Tatsuki and Isogawa, Mariko and Sakurada, Kuniharu and Sugimoto, Maki},
  booktitle = {SIGGRAPH Asia 2024 Posters},
  articleno = {53},
  numpages = {2},
  year = {2024},
  doi = {10.1145/3681756.3697916}
}

@inproceedings{cai2023utm,
  title     = {Robust Human Detection under Visual Degradation via Thermal and mmWave Radar Fusion},
  author    = {Cai, Kaiwen and Xia, Qiyue and Li, Peize and Stankovic, John and Lu, Chris Xiaoxuan},
  booktitle = {Proceedings of the International Conference on Embedded Wireless Systems and Networks (EWSN)},
  year      = {2023}
}

@inproceedings{choi2025mvdopplerpose,
  title     = {{MVDoppler-Pose}: Multi-Modal Multi-View mmWave Sensing for Long-Distance Self-Occluded Human Walking Pose Estimation},
  author    = {Choi, Jaeho and Hor, Soheil and Yang, Shubo and Arbabian, Amin},
  booktitle = {Proceedings of the IEEE/CVF Conference on Computer Vision and Pattern Recognition (CVPR)},
  year      = {2025}
}

@article{ren_gopose_2022,
  title = {{GoPose}: {3D} {Human} {Pose} {Estimation} {Using} {WiFi}},
  volume = {6},
  issn = {2474-9567},
  shorttitle = {{GoPose}},
  url = {https://dl.acm.org/doi/10.1145/3534605},
  doi = {10.1145/3534605},
  language = {en},
  number = {2},
  urldate = {2026-11-15},
  journal = {Proceedings of the ACM on Interactive, Mobile, Wearable and Ubiquitous Technologies},
  author = {Ren, Yili and Wang, Zi and Wang, Yichao and Tan, Sheng and Chen, Yingying and Yang, Jie},
  month = jul,
  year = {2022},
  pages = {1--25},
}

@article{adib_capturing_2015,
  title = {Capturing the human figure through a wall},
  volume = {34},
  issn = {0730-0301, 1557-7368},
  url = {https://dl.acm.org/doi/10.1145/2816795.2818072},
  doi = {10.1145/2816795.2818072},
  language = {en},
  number = {6},
  urldate = {2026-11-15},
  journal = {ACM Transactions on Graphics},
  author = {Adib, Fadel and Hsu, Chen-Yu and Mao, Hongzi and Katabi, Dina and Durand, Frédo},
  month = nov,
  year = {2015},
  pages = {1--13},
}

@inproceedings{zhang_mupose_2026,
  address = {University of Cambridge Cambridge United Kingdom},
  title = {{MuPose}: {Breaking} the {Scalability} {Barrier} of {mmWave} {Multi}-{User} {Pose} {Estimation} in the {Wild}},
  isbn = {979-8-4007-2027-7},
  shorttitle = {{MuPose}},
  url = {https://dl.acm.org/doi/10.1145/3745756.3809227},
  doi = {10.1145/3745756.3809227},
  language = {en},
  urldate = {2026-08-24},
  booktitle = {Proceedings of the 24th {Annual} {International} {Conference} on {Mobile} {Systems}, {Applications} and {Services}},
  publisher = {ACM},
  author = {Zhang, Duo and Yin, Zhehui and Zhang, Xusheng and Wang, Junzhe and Yang, Hongliu and Yao, Zhiyun and Fan, Zizhou and Li, Wenwei and Zhang, Daqing},
  month = jun,
  year = {2026},
  pages = {622--636},
}

@inproceedings{kong2022m3track,
  title={m3track: mmwave-based multi-user 3d posture tracking},
  author={Kong, Hao and Xu, Xiangyu and Yu, Jiadi and Chen, Qilin and Ma, Chenguang and Chen, Yingying and Chen, Yi-Chao and Kong, Linghe},
  booktitle={Proceedings of the 20th Annual International Conference on Mobile Systems, Applications and Services},
  pages={491--503},
  year={2022}
}

@inproceedings{zhang_pathclean_2026,
  address = {Austin TX USA},
  title = {{PathClean}: {Breaking} the {Multipath} {Barrier} for {High}-{Fidelity} {mmWave} {Human} {Point} {Clouds}},
  isbn = {979-8-4007-2505-0},
  shorttitle = {{PathClean}},
  url = {https://dl.acm.org/doi/10.1145/3795866.3844476},
  doi = {10.1145/3795866.3844476},
  language = {en},
  urldate = {2026-11-15},
  booktitle = {Proceedings of the 32nd {Annual} {International} {Conference} on {Mobile} {Computing} and {Networking}},
  publisher = {ACM},
  author = {Zhang, Duo and Yin, Zhehui and Zhang, Xusheng and Yang, Hongliu and Wang, Junzhe and Chang, Zhaoxin and Zhang, Fusang and Zhang, Daqing},
  month = oct,
  year = {2026},
  pages = {1139--1155},
}

@article{cao_occmesh_2025,
  title = {{OccMesh}: {Occlusion}-aware {Multi}-user {3D} {Human} {Mesh} {Reconstruction} {Using} {mmWave} {Signals}},
  volume = {9},
  issn = {2474-9567},
  shorttitle = {{OccMesh}},
  url = {https://dl.acm.org/doi/10.1145/3749514},
  doi = {10.1145/3749514},
  language = {en},
  number = {3},
  urldate = {2026-08-24},
  journal = {Proceedings of the ACM on Interactive, Mobile, Wearable and Ubiquitous Technologies},
  author = {Cao, Haoran and Yu, Jiadi and Kong, Hao and Chen, Yi-Chao and Zhu, Yanmin and Kong, Linghe and Tang, Feilong},
  month = sep,
  year = {2025},
  pages = {1--22},
}

@article{peng_expanding_nodate,
  title={Expanding mmWave Datasets for Human Pose Estimation with Unlabeled Data and LiDAR Datasets},
  author={Peng, Zhuoxuan and Zhu, Boan and Zhang, Xingjian and Li, Wenying and Chan, S-H Gary},
  journal={arXiv preprint arXiv:2603.14507},
  year={2026}
}

@article{yang_spatial_2025,
  title = {From {Spatial} {Domain} to {Temporal} {Domain}: {Unleashing} the {Capability} of {CFAR} for {mmWave} {Point} {Cloud} {Generation}},
  volume = {9},
  issn = {2474-9567},
  shorttitle = {From {Spatial} {Domain} to {Temporal} {Domain}},
  url = {https://dl.acm.org/doi/10.1145/3729465},
  doi = {10.1145/3729465},
  language = {en},
  number = {2},
  urldate = {2026-01-14},
  journal = {Proceedings of the ACM on Interactive, Mobile, Wearable and Ubiquitous Technologies},
  author = {Yang, Hongliu and Zhang, Duo and Zhang, Xusheng and Xiong, Jie and Fan, Zizhou and Ning, Wanru and Chen, Weiyan and Zhang, Fusang and Han, Zijun and Zhang, Daqing},
  month = jun,
  year = {2025},
  pages = {1--29},
}

@inproceedings{xue_m4_2022,
  address = {Boston Massachusetts},
  title = {M$^{\textrm{4}}$ esh: {mmWave}-{Based} {3D} {Human} {Mesh} {Construction} for {Multiple} {Subjects}},
  isbn = {978-1-4503-9886-2},
  shorttitle = {M$^{\textrm{4}}$ esh},
  url = {https://dl.acm.org/doi/10.1145/3560905.3568545},
  doi = {10.1145/3560905.3568545},
  language = {en},
  urldate = {2026-08-24},
  booktitle = {Proceedings of the 20th {ACM} {Conference} on {Embedded} {Networked} {Sensor} {Systems}},
  publisher = {ACM},
  author = {Xue, Hongfei and Cao, Qiming and Ju, Yan and Hu, Haochen and Wang, Haoyu and Zhang, Aidong and Su, Lu},
  month = nov,
  year = {2022},
  pages = {391--406},
}

@inproceedings{yang_omnipc_2026,
  address = {Saint Malo France},
  title = {{OmniPC}: {A} {Generalizable} {Point} {Cloud} {Generation} {Pipeline} for {mmWave} {Radar}},
  isbn = {979-8-4007-2309-4},
  shorttitle = {{OmniPC}},
  url = {https://dl.acm.org/doi/10.1145/3774906.3802797},
  doi = {10.1145/3774906.3802797},
  language = {en},
  urldate = {2026-08-27},
  booktitle = {Proceedings of the 2026 {ACM}/{IEEE} {International} {Conference} on {Embedded} {Artificial} {Intelligence} and {Sensing} {Systems}},
  publisher = {ACM},
  author = {Yang, Hongliu and Fan, Zizhou and Xiong, Jie and Zhang, Duo and Zhang, Xusheng and Han, Zijun and Zhang, Fusang and Zhang, Daqing},
  month = may,
  year = {2026},
  pages = {291--304},
}

@article{wang_human_2023,
  title = {Human {Parsing} with {Joint} {Learning} for {Dynamic} {mmWave} {Radar} {Point} {Cloud}},
  volume = {7},
  issn = {2474-9567},
  url = {https://dl.acm.org/doi/10.1145/3580779},
  doi = {10.1145/3580779},
  language = {en},
  number = {1},
  urldate = {2024-05-17},
  journal = {Proceedings of the ACM on Interactive, Mobile, Wearable and Ubiquitous Technologies},
  author = {Wang, Shuai and Cao, Dongjiang and Liu, Ruofeng and Jiang, Wenchao and Yao, Tianshun and Lu, Chris Xiaoxuan},
  month = mar,
  year = {2023},
  pages = {1--22},
}

@article{kim2024multipath,
  title={Multipath suppression using 2-D angle analysis based on the MIMO FMCW radar},
  author={Kim, Jihye and Jung, Jaehoon and Lim, Sohee and Kim, Seong-Cheol},
  journal={IEEE Transactions on Instrumentation and Measurement},
  volume={74},
  pages={1--12},
  year={2024},
  publisher={IEEE}
}

@article{chen2023environment,
  title={Environment-aware multi-person tracking in indoor environments with mmWave radars},
  author={Chen, Weiyan and Yang, Hongliu and Bi, Xiaoyang and Zheng, Rong and Zhang, Fusang and Bao, Peng and Chang, Zhaoxin and Ma, Xujun and Zhang, Daqing},
  journal={Proceedings of the ACM on Interactive, Mobile, Wearable and Ubiquitous Technologies},
  volume={7},
  number={3},
  pages={1--29},
  year={2023},
  publisher={ACM, New York, NY, USA}
}

@article{yang_bridging_2026,
  title = {Bridging the {Resolution} {Gap}: {Cost}-{Effective} {Human} {Point} {Cloud} {Generation} via {Low}-{Bandwidth} {mmWave} {Radar}},
  volume = {10},
  issn = {2474-9567},
  shorttitle = {Bridging the {Resolution} {Gap}},
  url = {https://dl.acm.org/doi/10.1145/3810223},
  doi = {10.1145/3810223},
  language = {en},
  number = {2},
  urldate = {2026-08-27},
  journal = {Proceedings of the ACM on Interactive, Mobile, Wearable and Ubiquitous Technologies},
  author = {Yang, Hongliu and Fan, Zizhou and Wang, Yueyang and Xiong, Jie and Zhang, Duo and Zhang, Xusheng and Wang, Junzhe and Zhang, Fusang and Zhang, Daqing},
  month = jun,
  year = {2026},
  pages = {1--26},
}

@inproceedings{yang_mmdear_2025,
  title={mmDEAR: mmWave point cloud density enhancement for accurate human body reconstruction},
  author={Yang, Jiarui and Xia, Songpengcheng and Lai, Zengyuan and Sun, Lan and Wu, Qi and Yu, Wenxian and Pei, Ling},
  booktitle={2025 IEEE International Conference on Robotics and Automation (ICRA)},
  pages={11227--11233},
  year={2025},
  organization={IEEE}
}

@inproceedings{yang2023mmfi,
  title = {{MM-Fi}: Multi-Modal Non-Intrusive 4D Human Dataset for Versatile Wireless Sensing},
  author = {Yang, Jianfei and Huang, He and Zhou, Yunjiao and Chen, Xinyan and Xu, Yuecong and Yuan, Shenghai and Zou, Han and Lu, Chris Xiaoxuan and Xie, Lihua},
  booktitle = {Advances in Neural Information Processing Systems},
  volume = {36},
  year = {2023},
  url = {https://proceedings.neurips.cc/paper_files/paper/2023/hash/3baf7a39d07e9f4f1e258a412df94521-Abstract-Datasets_and_Benchmarks.html}
}

@inproceedings{ho2025rtpose,
  title = {{RT-Pose}: A 4D Radar Tensor-Based 3D Human Pose Estimation and Localization Benchmark},
  author = {Ho, Yuan-Hao and Cheng, Jen-Hao and Kuan, Sheng Yao and Jiang, Zhongyu and Chai, Wenhao and Huang, Hsiang-Wei and Lin, Chih-Lung and Hwang, Jenq-Neng},
  booktitle = {Computer Vision -- ECCV 2024},
  series = {Lecture Notes in Computer Science},
  volume = {15121},
  pages = {107--125},
  publisher = {Springer},
  year = {2025},
  doi = {10.1007/978-3-031-73036-8_7}
}

@inproceedings{kurihara2021boundary,
  title = {Boundary Guided Image Translation for Pose Estimation from Ultra-Low Resolution Thermal Sensor},
  author = {Kurihara, Kohei and Wang, Tianren and Zhang, Teng and Lovell, Brian C.},
  booktitle = {2020 25th International Conference on Pattern Recognition (ICPR)},
  pages = {9--15},
  year = {2021},
  doi = {10.1109/ICPR48806.2021.9412250}
}

@article{zhao2023nighttime,
  title = {Nighttime Pedestrian Detection Based on a Fusion of Visual Information and Millimeter-Wave Radar},
  author = {Zhao, Wei and Wang, Tingting and Tan, Ao and Ren, Congcong},
  journal = {IEEE Access},
  volume = {11},
  pages = {68439--68451},
  year = {2023},
  doi = {10.1109/ACCESS.2023.3291398}
}

@article{li2024riders,
  title = {{RIDERS}: Radar-Infrared Depth Estimation for Robust Sensing},
  author = {Li, Han and Ma, Yukai and Huang, Yuehao and Gu, Yaqing and Xu, Weihua and Liu, Yong and Zuo, Xingxing},
  journal = {IEEE Transactions on Intelligent Transportation Systems},
  volume = {25},
  number = {11},
  pages = {18764--18778},
  year = {2024},
  doi = {10.1109/TITS.2024.3432996}
}

@article{rintahomi2021howlow,
  title = {How Low Can You Go? Performance Trade-offs in Low-Resolution Thermal Sensors for Occupancy Detection: A Systematic Evaluation},
  author = {Rinta-Homi, Mikko and Hossein Motlagh, Naser and Zuniga, Agustin and Flores, Huber and Nurmi, Petteri},
  journal = {Proceedings of the ACM on Interactive, Mobile, Wearable and Ubiquitous Technologies},
  volume = {5},
  number = {3},
  articleno = {126},
  numpages = {22},
  year = {2021},
  doi = {10.1145/3478104}
}
}
\end{document}